\documentclass[lettersize,journal]{IEEEtran}

\usepackage{amsmath,amssymb,amsfonts}
\usepackage{algorithmic}
\usepackage{algorithm}
\usepackage{array}
\usepackage[caption=false,font=normalsize,labelfont=sf,textfont=sf]{subfig}
\usepackage{textcomp}
\def\BibTeX{{\rm B\kern-.05em{\sc i\kern-.025em b}\kern-.08em
    T\kern-.1667em\lower.7ex\hbox{E}\kern-.125emX}}
\usepackage{stfloats}
\usepackage{url}
\usepackage{verbatim}
\usepackage{graphicx}
\usepackage{cite}
\usepackage{makecell}
\usepackage{dsfont}
\usepackage{bbding}
\usepackage{multirow}
\usepackage{bm}
\usepackage{threeparttable}
\usepackage{soul}
\usepackage{color}

\begin{document}
\title{Addressing Domain Shift via Knowledge Space Sharing for Generalized Zero-Shot Industrial Fault Diagnosis}
\author{Jiancheng~Zhao, Jiaqi~Yue, Liangjun~Feng, Chunhui~Zhao,~\IEEEmembership{Senior~Member,~IEEE}, and Jinliang~Ding, \IEEEmembership{Senior~Member,~IEEE}
\thanks{This work is supported by the National Science Fund for Distinguished Young Scholars (No. 62125306), the Guangdong Basic and Applied Basic Research Foundation (2022A1515240003)}
\thanks{Jiancheng Zhao, Jiaqi Yue, Liangjun Feng and Chunhui Zhao are with the College of Control Science and Engineering, Zhejiang University, Hangzhou 310027, China. (Email: zhaojiancheng@zju.edu.cn, 12232051@zju.edu.cn, liangjunfeng@zju.edu.cn, chhzhao@zju.edu.cn) }
\thanks{Jinliang Ding is with the State Key Laboratory of Synthetical Automation for Process Industries, Northeastern University, Shenyang 110819, China (Email: jlding@mail.neu.edu.cn)}
\thanks{The corresponding author is Chunhui Zhao.}
\thanks{This work has been submitted to the IEEE for possible publication. Copyright may be transferred without notice, after which this vesion may no longer be accessible.}}

\markboth{IEEE TRANSACTIONS ON AUTOMATION SCIENCE AND ENGINEERING}%
{Shell \MakeLowercase{\textit{et al.}}: A Sample Article Using IEEEtran.cls for IEEE Journals}


\maketitle
\begin{abstract}
Fault diagnosis is a critical aspect of industrial safety, and supervised industrial fault diagnosis has been extensively researched. However, obtaining fault samples of all categories for model training can be challenging due to cost and safety concerns. As a result, the generalized zero-shot industrial fault diagnosis has gained attention as it aims to diagnose both seen and unseen faults. Nevertheless, the lack of unseen fault data for training poses a challenging domain shift problem (DSP), where unseen faults are often identified as seen faults. In this article, we propose a knowledge space sharing (KSS) model to address the DSP in the generalized zero-shot industrial fault diagnosis task. The KSS model includes a generation mechanism (KSS-G) and a discrimination mechanism (KSS-D). KSS-G generates samples for rare faults by recombining transferable attribute features extracted from seen samples under the guidance of auxiliary knowledge. KSS-D is trained in a supervised way with the help of generated samples, which aims to address the DSP by modeling seen categories in the knowledge space. KSS-D avoids misclassifying rare faults as seen faults and identifies seen fault samples. We conduct generalized zero-shot diagnosis experiments on the benchmark Tennessee-Eastman process, and our results show that our approach outperforms state-of-the-art methods for the generalized zero-shot industrial fault diagnosis problem. 
\end{abstract}
\def\abstractname{Note to Practitioners}
\begin{abstract}
This paper is motivated by the difficulty of fault diagnosis in practical industrial scenarios caused by the lack of fault samples for training supervised diagnosis models. The focus of this study is to develop a generalized zero-shot industrial fault diagnosis method, which can diagnose both faults with sufficient training samples and faults without training samples with the help of expert knowledge of these faults. Considering the generalized zero-shot diagnosis models often confuse unseen faults as seen faults, a knowledge space sharing model is given to address this problem. It contains a generator to provide generated samples for rare faults and a discriminator to differentiate unseen faults from seen faults and classify them in the knowledge space. The experiments conducted on the benchmark Tennessee-Eastman process suggest that this approach is feasible. In future research, we will consider incremental learning for the generalized zero-shot diagnosis method.

\end{abstract}
\def\abstractname{abstract}

\section{Introduction}
\label{sec:introduction}
\IEEEPARstart{F}{ault} diagnosis is a crucial task for maintaining industrial equipment, and various data-driven approaches have been proposed to enhance the ability to diagnose complex and diverse faults\cite{qin2012survey,zhao2022perspectives}. Traditionally, data-driven fault diagnosis has been modeled as a supervised classification task, where sufficient samples of target faults are collected for modeling and training. Many supervised data-driven diagnosis (SD) methods have been proposed, such as support vector machine (SVM)\cite{yin2016recent}, dictionary learning\cite{9963791}, and deep neural network\cite{chai2020fine}, to identify faults. These methods have been successful when enough labeled data of faults can be collected. However, in practical industrial scenarios, plants often operate normally for extended periods, resulting in fewer fault data being collected compared to normal data. Consequently, the actual distribution of such types of faults tends to be long-tailed. Additionally, conducting fault simulation experiments on real equipment is not always feasible due to cost and safety concerns\cite{sun2018deep}. Therefore, it is common for industrial fault diagnosis scenarios to have few or no available data of target faults for training\cite{9881217}. The scarcity of samples poses a constraint on the performance of SD.

The lack of samples is a significant problem in various fields. In computer vision, attributes like colors, shapes, and surroundings have been summarized from images. These attributes serve as a bridge between seen and unseen categories, enabling zero-shot classification. Lampert et al. \cite{lampert2009learning} proposed zero-shot learning (ZSL) to address the challenge of zero-shot image classification, where no images of testing categories are available for training. To transfer knowledge from training categories to target testing categories, a group of attribute predictors was trained using training samples and their corresponding attributes, so that testing samples were identified by their predicted attributes. This approach is termed direct attribute prediction (DAP). Different from the DAP, Akata et al. \cite{akata2015label,akata2015evaluation} proposed compatibility functions to evaluate the compatibility between samples and attributes. Unseen samples were identified by finding the labels that yielded the highest compatibility score. Additionally, Romera-Paredes et al. \cite{romera2015embarrassingly} proposed an embarrassingly simple approach to zero-shot learning (ESZSL), which modeled the relationships between features, attributes, and classes as a two-layer linear network.

For zero-shot image classification, attributes like shapes and colors are proven to be effective. However, in industrial scenarios, these visual attributes are not suited and more professional auxiliary knowledge is required for fault diagnosis. Therefore, Feng et al. \cite{feng2020fault} first proposed the zero-shot diagnosis (ZSD) problem and defined professional fault description as auxiliary knowledge, which provided the basis for later works. They improved the DAP to identify the attributes of unseen faults. The attributes summarized from the fault description can be damaged components, fault causes and so on. Based on their work, more realistic task setting\cite{huang2021fault} and generative-based methods\cite{zhuo2021auxiliary, 10040605} were introduced in the follow-up work. According to the difference in the test stage, ZSD can be divided into traditional ZSD (TZSD) and generalized ZSD (GZSD)\cite{pourpanah2022review,feng2020transfer}. TZSD only distinguishes samples of unseen categories in the test stage with the help of auxiliary knowledge. To be more realistic, GZSD is required to diagnose both seen and unseen faults at the same time, so GZSD is more challenging compared to others. Also, different from TZSD, GZSD faces the domain shift problem\cite{pourpanah2022review}, where unseen faults tend to be classified as seen faults due to the lack of unseen fault samples for training. 

Existing ZSD methods can be categorized into two types: embedding-based and generative-based. Embedding-based methods \cite{feng2020fault} learn a mapping between samples and auxiliary knowledge to identify unseen faults. For instance, Xu et al. \cite{xu2022zero} utilized a convolutional neural network as a feature extractor and embedded the semantic features of faults into the visual space through a two-layer fully connected network to accomplish zero-shot compound fault diagnosis. However, embedding-based methods are prone to the DSP due to the overfitting of seen faults. On the other hand, generative-based methods transform ZSD into supervised learning by generating samples of unseen categories based on knowledge. This approach can alleviate the DSP by generating fake samples for training the classifier. For example, Huang et al. \cite{huang2021fault} employed a conditional variational auto-encoder (CVAE), while Zhuo et al. \cite{zhuo2021auxiliary} proposed a generative adversarial model using fault attributes (FAGAN) to generate samples of unseen categories by utilizing attribute vectors as initial conditions for generating specific categories. While the generative-based approach has been demonstrated to be effective, it remains unclear how knowledge guides the generation process. In addition, nearly all generative models include a noise component to provide diversity in generating process, which may reduce the quality of generated samples\cite{arjovsky2017wasserstein}.

Although different methods build the relationship between knowledge and features in different ways, DSP is a common and knotty problem for GZSD as a result of the lack of unseen fault samples for model training. To overcome the DSP, several approaches have been proposed, such as transductive-based methods \cite{cheraghian2020transductive}, calibrated stacking \cite{chao2016empirical}, gating mechanism \cite{socher2013zero}, and so on. The transductive-based methods are not suitable for real-time diagnosis, because they require online accumulation of samples. Methods based on calibrated stacking use a manually set calibration factor to reduce the confidence of assigning test samples to the seen categories, which is not adaptive and universal. The gating mechanism distinguishes unseen samples from seen ones by a binary classifier. However, the existing gating mechanism methods are mostly done in the feature space, and therefore, training the gate is challenging since real features of the unseen classes are unavailable\cite{pourpanah2022review}. Furthermore, the performance of the gate degrades significantly when available fault samples are insufficient, which is quite typical in the real industrial environment.

To address the aforementioned DSP for generalized zero-shot industrial fault diagnosis, we propose a \textbf{k}nowledge \textbf{s}pace \textbf{s}haring (KSS) model, which contains a generation mechanism (KSS-G) and a discrimination mechanism (KSS-D). KSS-G overcomes the lack of samples by combining different attribute features under the guidance of auxiliary knowledge, which avoids the decrease in generated sample quality caused by introducing noise. Compared to being used as condition vectors in CVAE or CGAN, auxiliary knowledge can interpretably participate in the generation process. KSS-D distinguishes test samples into seen or unseen faults by modeling the projection distribution of seen faults in the knowledge space. It also performs coarse-to-fine classification for seen faults. With the fake samples from KSS-G, KSS-D works more accurately. In this paper, KSS refers to the introduction of knowledge not only to the generator (KSS-G) but also to the discrimination mechanism (KSS-D). This approach is beneficial in addressing DSP as the discrimination mechanism receives additional information on both seen and unseen categories. The contributions of this paper are summarized below.
\begin{enumerate}
\item{We propose a knowledge space sharing paradigm to address the DSP in GZSD tasks. Our approach introduces knowledge not only to the generator but also to the discrimination mechanism. In contrast, existing GZSD methods only introduce knowledge to generators, which can result in poor performance of the discrimination mechanism due to the lack of unseen faults.}
\item{We propose a novel generator (KSS-G) that synthesizes samples without introducing extra noise through knowledge-supervised attribute feature extraction and reorganization. Compared to being used as the input condition in CVAE or CGAN, knowledge is interpretably involved in the generation process of KSS-G.}
\item{We propose a novel discrimination mechanism (KSS-D) to differentiate between the categories of seen and unseen. It is the first gating mechanism that is based on knowledge space and mitigates the DSP by modeling each seen category distribution in the knowledge space.}
\end{enumerate}

This paper is structured as follows. Section \ref{sec:problem} discusses the problem formulation and the motivation behind this research. Section \ref{sec:method} presents the proposed KSS model, which consists of KSS-G and KSS-D. Section \ref{sec:case} evaluates the performance of the proposed model on the GZSD task using the benchmark Tennessee-Eastman process. This section also compares the proposed model with state-of-the-art methods, visualizes the results, conducts an ablation study, and contrasts it with generalized few-shot learning. Section \ref{sec:con} concludes this paper.

\section{Problem Formulation and Motivation}
\label{sec:problem}
\subsection{Problem Formulation}
For GZSD, the accessible fault categories for training are denoted as $S=\{s_1,\ldots, s_p\}$, where $p$ is the number of seen faults. The fault categories that are unavailable for training are denoted as $U=\{u_1,\ldots, u_q\}$, where $q$ is the number of unseen faults, and $S \cap U = \emptyset$. Samples of $S$ are denoted as $\{ \bm{X}^s \in \mathbb{R}^{N^s \times d}, \bm{Y}^s \in \mathbb{R}^{N^s} \}$, where $N^s$ is the number of training samples, and $d$ is the dimension of the features. It is convenient to use one-hot coding to represent the fault attributes for seen or unseen faults labeled by experts. The description of the fault category $j$ is denoted as $\bm{a}^j \in \mathbb{R}^{M}, j = 1,2, ..., L$ where $M$ is the number of attributes and $L$ is the number of fault categories. Each element in $\bm{a}^j$ is either 1 or 0, depending on whether the attribute exists or not in the fault category $j$. The fault category-attributes matrix for all of the $L$ fault categories is denoted as $\bm{A} = [\bm{A}^s, \bm{A}^u] \in \mathbb{R}^{L \times M}$, where $\bm{A}^s \in \mathbb{R}^{p \times M}$ and $\bm{A}^u \in \mathbb{R}^{q \times M}$ are the attribute matrix for the $S$, $U$, respectively. It is worth mentioning that both the attribute description matrix $\bm{A}^s$ and $\bm{A}^u$ are available in the training and testing stage since descriptions are class-level rather than sample-level. We merge the training label $\bm{Y}^s$ with the fault category-attributes matrix $\bm{A}^s$ to obtain the training attribute label $\bm{Z}^s \in \mathbb{R}^{N^s \times M}$. Thus, the training set can be denoted by $\mathbb{D}_{tr}=\{ (\bm{x}^{i},y^{i},\bm{z}^{i})|\bm{x}^{i} \in \bm{X}^s,y^{i} \in \bm{Y}^s, \bm{z}^{i} \in \bm{Z}^s \}$. In the testing stage, the testing set can be denoted by $\mathbb{D}_{ts}=\{ (\bm{x}^{i}, y^{i})|\bm{x}^{i} \in \bm{X}^t \in \mathbb{R}^{N^t \times d},y^{i} \in \bm{Y}^t \in \mathbb{R}^{N^t} \}$, where $\bm{X}^t$ contains test samples of both seen and unseen categories and $\bm{Y}^t$ contains target class labels. We aim to learn a classifier $f$, which can be formulated as
\begin{equation}
\begin{aligned}
& Min \quad {\rm ClsLoss} (\bm{Y}^t,\hat{\bm{Y}}^t), \\
& and \quad  \hat{\bm{Y}}^t=f (\bm{X}^s,\bm{Y}^s,\bm{A}^s,\bm{A}^u|\bm{X}^t),
\end{aligned}
\end{equation}
where ${\rm ClsLoss}$ denotes an arbitrary classification loss function. 

\subsection{Comments and Motivation}
Some initial studies on ZSD have been proposed in recent years. We have summarized the core of these studies into three main issues, which we refer to as the "3H" issues. The existing studies on these 3H issues are presented in Table \ref{tab:model_com_detail}. 
\begin{table}[!t]
\renewcommand{\arraystretch}{1}
\setlength\tabcolsep{0.3pt}
\caption{Comparison of Different Zero-Shot Diagnosis Methods}
\label{tab:model_com_detail}
\centering
\begin{threeparttable}
\begin{tabular}{cccc}
\hline
\hline
Main models & \emph{H1}\tnote{1} & \emph{H2} & \emph{H3} \\
\hline 
SPCA+DAP  \cite{feng2020fault} & Embedding & Embedding & KNN (u)  \\
CVAE \cite{huang2021fault} & \makecell{Generating , \\gating , embedding}&\makecell{Generating data,\\embedding} & RF (s), KNN (u) \\
FAGAN \cite{zhuo2021auxiliary} & Generating  & Generating data & KNN (s, u) \tnote{2} \\
CADAE \cite{lv2020hybrid} & \makecell{Generating ,\\Reconstruction } & Generating data & DNN (u) \\
LDS-IFD \cite{xing2022label} & \makecell{Generating , \\gating } & \makecell{Generating prototypes,\\embedding} & \makecell{Softmax (s) ,\\KNN (u)} \\
ZLCFDM \cite{xu2022zero} & Embedding & \makecell{Generating prototypes,\\embedding} & KNN (u) \\
GAN \cite{xu2021generative} & Generating  & Generating data & KNN (u) \\
CGAN \cite{chen2019adversarial},\cite{pan2020deep} & Generating  & \textbackslash{} & DNN (u) \\
CSA \cite{gao2020zero} & \textbackslash{} & \textbackslash{} & Softmax (u) \\
SECM \cite{gupta2020unseen} & Embedding & Embedding & Compatibility (u) \\
\hline
\hline 
\end{tabular}

\begin{tablenotes}
\footnotesize
\item[1] Since feature extraction is a common step in most methods, it has been omitted for clarity in \emph{H1}.
\item[2] The letters 's' and 'u' indicate whether the classifier is used for seen or unseen faults, respectively. Only GZSL methods need the classifier for seen categories.
\end{tablenotes}

\end{threeparttable}
\end{table}

\subsubsection{H1} How to use the data of seen faults?

The foundation of ZSD lies in utilizing seen samples without overfitting the seen faults. To achieve this, data from seen categories are often employed to train modules that are shared between both seen and unseen categories. These modules include feature extractors, embedding modules, generators, gating modules \cite{huang2021fault}, and reconstruction modules \cite{lv2020hybrid}.

\subsubsection{H2} How to establish a relationship between the knowledge space and the feature space?

The main challenge for ZSD is to establish the relationship between knowledge and feature. There are two main types of ZSD methods: embedding-based and generating-based, which establish this relationship in different ways. Embedding-based methods create a mapping between knowledge and feature, but they are not suitable for GZSD due to overfitting. On the other hand, generating methods can be divided into two types: data generation (generative-based methods) and category prototype generation. Generative-based methods convert the ZSL paradigm into a supervised paradigm by generating unseen class data, while some methods generate unseen category prototypes as anchors so that KNN-based classifiers can be used for unseen categories. The use of fake samples from generative-based methods helps to avoid overfitting to seen samples, which is particularly meaningful for DSP.

\subsubsection{H3} How to implement classification?

The choice of classifiers depends on the nature of the adopted methods. Metric-based KNN classifiers are typically favored for techniques that involve embedding or generating category prototypes. Other classifiers, such as compatibility-based classifiers\cite{gupta2020unseen}, K-Nearest Neighbor (KNN), Deep Neural Network (DNN), Random Forest (RF), Naive Bayes (NB), are utilized to classify. 

Our work aims to extract transferable representations from seen data in an interpretable manner, for generating both seen and unseen faults. To achieve this, we propose KSS-G, a novel generative-based method that extracts and restructures transferable attribute representations with the supervision of auxiliary knowledge. This approach effectively alleviates DSP. Additionally, we introduce KSS-D to distinguish between seen and unseen faults by modeling the distribution of seen faults in the knowledge space. To the best of our knowledge, it is the first gating mechanism that is based on knowledge space. Both KSS-G and KSS-D transfer information from seen to unseen faults under the guidance of auxiliary knowledge.

\section{Methodology}
\label{sec:method}
\begin{figure}[!t]
\centering
\includegraphics[width=1 \linewidth]{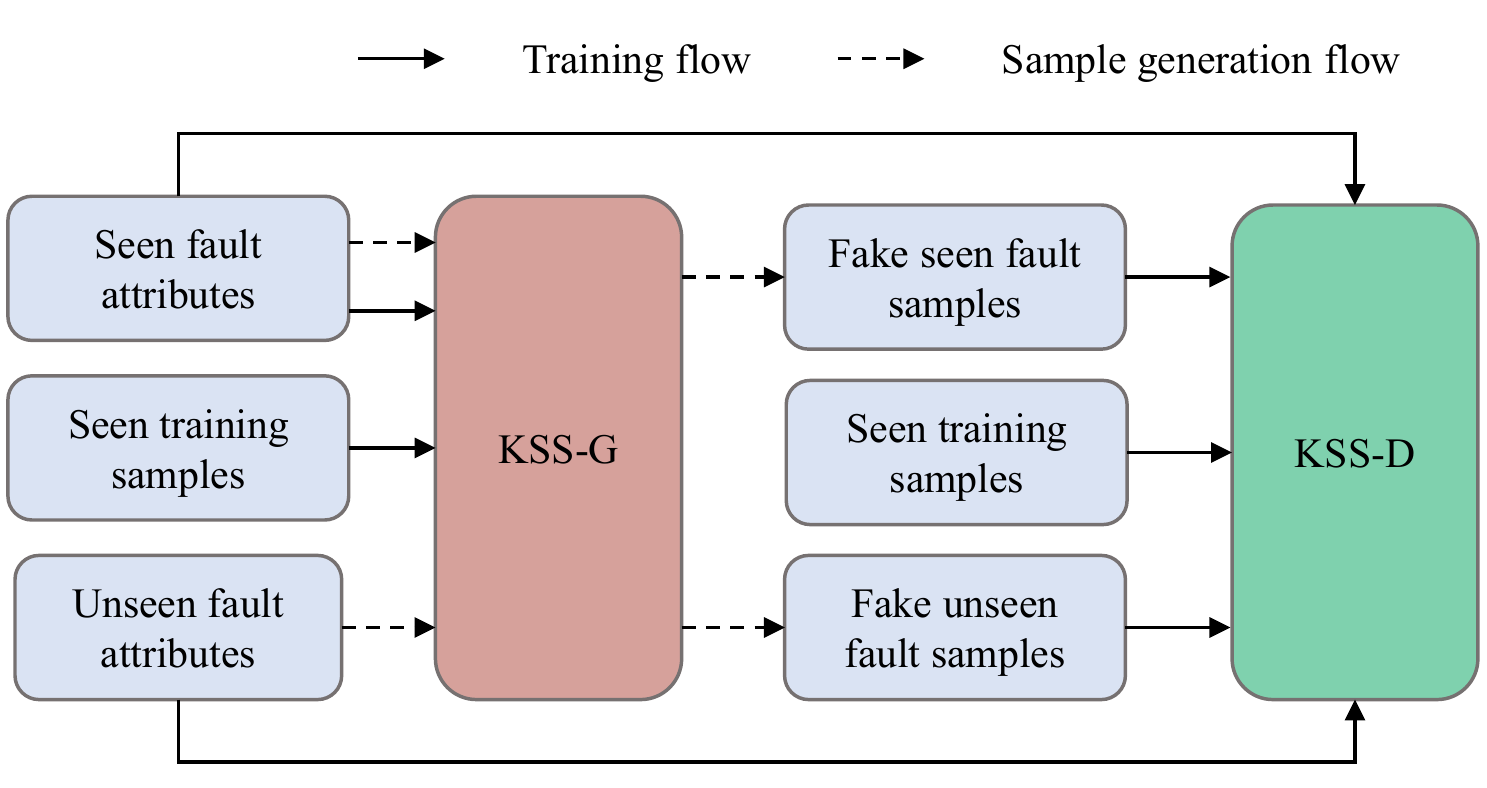}
\caption{The basic rationale of the proposed KSS model.}
\label{fig:basic}
\end{figure}

In this section, we will provide a detailed description of the proposed model. Firstly, we will introduce the basic rationale behind the KSS model. Subsequently, each component of the model will be demonstrated. Figure \ref{fig:basic} illustrates the fundamental principle of the KSS model. To begin with, a generator KSS-G is trained using training samples and attributes. Next, KSS-G is utilized to generate fake samples for both seen and unseen faults, which enables us to convert the zero-shot learning problem into a traditional supervised learning problem. Finally, we train KSS-D using real and fake samples to classify faults accurately for both seen and unseen categories. The training and testing process is as follows: Firstly, the aid-discriminator is pre-trained, consisting of a backbone and three branches: the multi-class classification branch, attribute recognition branch, and discrimination branch. In this step, we pre-train the aid-discriminator except for the aid-discrimination branch. Secondly, we train the generator (KSS-G). The style of the generative adversarial network (GAN) is made up of the reconstruction module and the aid-discriminator in KSS-G. During training KSS-G, samples of seen categories are generated. Finally, a KSS-D is trained to differentiate between test samples that belong to seen categories and those that do not by modeling the seen categories in the knowledge space. If a test sample belongs to a seen category, its specific label is identified; otherwise, a TZSD model trained based on real and fake samples is used to determine its category.

\subsection{Generating Samples via Knowledge Space Sharing}
\begin{figure*}[!t]
\centering
\includegraphics[width=0.75 \linewidth]{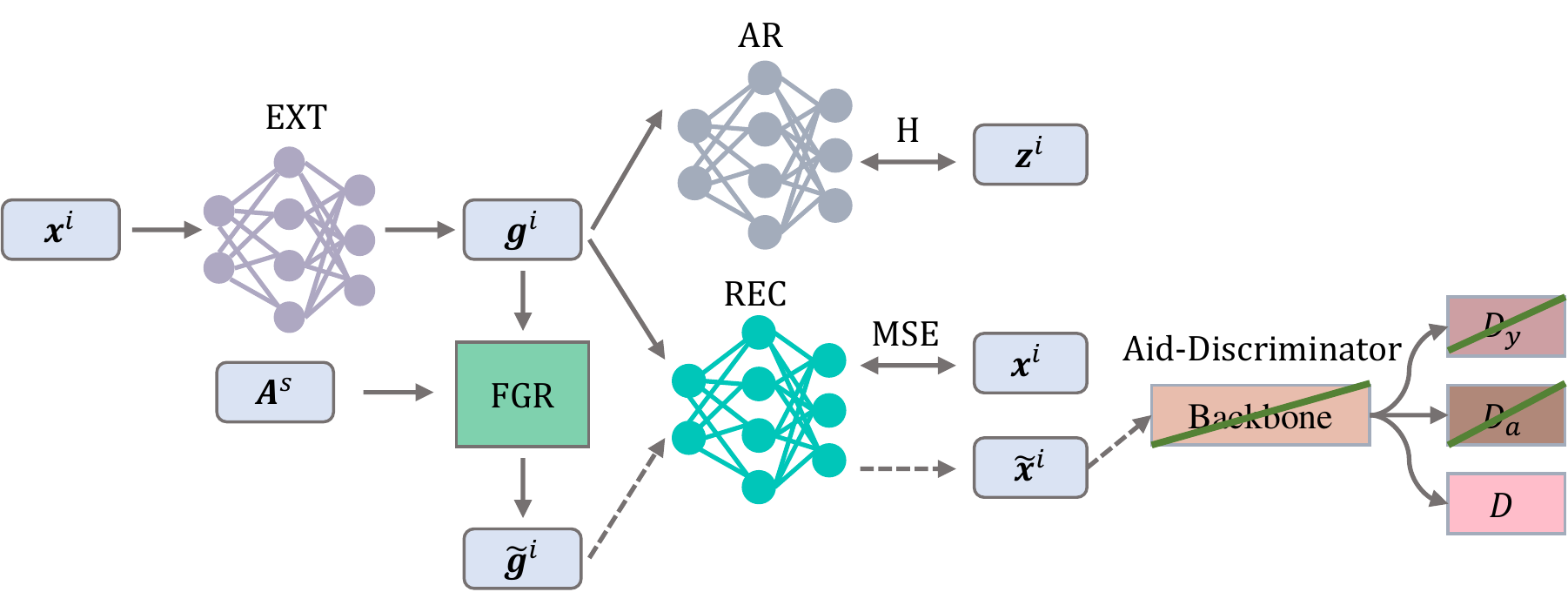}
\caption{The diagram of the generation module (KSS-G). In the KSS-G, the attribute feature $\bm{g}^s$ is first extracted by ${\rm EXT}$. Then ${\rm AR}$ recognizes the attributes of $\bm{g}^s$. The feature group reorganization (FGR) module reorganizes the features of seen samples so that the reorganized feature $\widetilde{\bm{g}}^s$ meets the needs of the target category. Finally, $\rm REC$ reconstructs train samples by $\bm{g}^s$ and generates fake samples by $\widetilde{\bm{g}}^s$. The aid-discriminator supervises the quality of fake samples using a zero-sum game (green lines mean frozen parameters). }
\label{fig:KSS-G}
\end{figure*}

The architecture of KSS-G is illustrated in Fig. \ref{fig:KSS-G}. KSS-G can be viewed as a mapping between knowledge and samples that have been trained on seen categories, which can generate fake samples of those unseen categories by utilizing the expertise of professionals on unseen categories. Unlike other generative-based methods that only introduce knowledge as a starting condition, KSS-G extracts and recombines features representing attributes from the original samples to generate fake samples, which involves the knowledge interpretably and avoids the decrease in generated sample quality caused by introducing extra noise terms. To achieve this, a group of feature extractors is used to extract features representing different attributes. Then, a group of attribute recognizers is trained with supervision to ensure that attribute information is contained. A feature group reorganization (FGR) operation is used to reorganize these feature groups according to the attribute labels of seen or unseen categories. Finally, a reconstruction module is used to transfer feature groups into fake samples. Additionally, an aid-discriminator is used to improve generation quality through adversarial training. KSS-G is denoted as $\Theta_{g}$ in the following.

\begin{figure*}[!t]
    \centering
    \includegraphics[width=0.8 \linewidth]{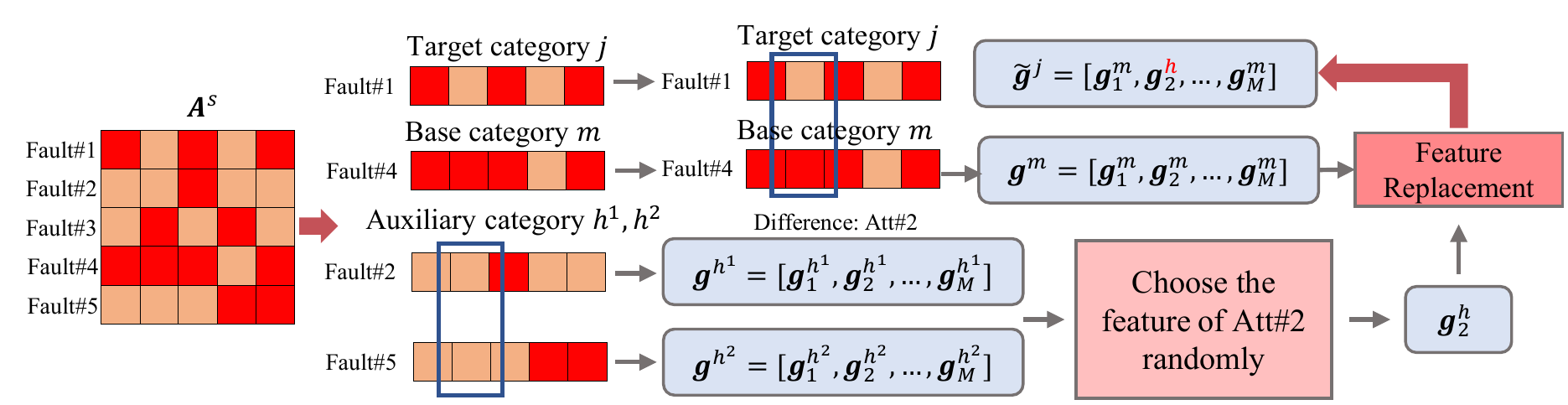}
    \caption{The diagram of the feature group reorganization (FGR). FGR operation recombines attribute features under the guidance of the auxiliary knowledge to generate fake samples. To illustrate, suppose that we intend to generate a fake sample of category$j$ (Fault\#1). Initially, we evaluate the L1-norm distance between $\bm{a}^j$ and each row of $\bm{A}^s$ except for the row $j$ to determine the seen category $m$ that best resembles category $j$. By comparing $\bm{a}^j$ and $\bm{a}^m$, we can find the attribute (ATT\#2) that is different between categories $j$ and $m$. Afterward, we randomly select a sample $h$ that has the same ATT\#2 as the category $j$, and utilize its feature of ATT\#2 to supplant the corresponding feature in the attribute feature group extracted from category $m$, thereby obtaining the feature utilized for generating fake samples of category $j$.}
    \label{fig:FGR}
\end{figure*}

\subsubsection{Aid-Discriminator for Training KSS-G}
As previously mentioned, the aid-discriminator consists of a backbone and three branches. We use real data to pre-train the backbone, the multiclass classification branch, and the attribute recognition branch by predicting the probabilities that input samples belong to each fault type and identifying attributes. The pre-trained portion of the aid-discriminator is referred to as $\Theta_{b}$. The loss function for pre-training the aid-discriminator is expressed as follows:
\begin{equation}
    \mathcal{L}_{AD} \!=\! \frac{1}{N^s}\!\sum_{i=1}^{N^s}\left [ {\rm H} (y^i,D_y (\bm{x}^i))\!+\!\frac{1}{M}\!\sum_{k=1}^{M}{\rm H} (z_{k}^i,D_a (\bm{x}^i)_{k})\right ],
\end{equation}
where $y^i$ denotes the fault type label corresponding to $\bm{x}^i$, $z_{k}^{i}$ is the $k$th value of $\bm{z}^i$, $D_y (\cdot)$ denotes multiclass classification, $D_a (\cdot)$ denotes attribute recognition, and ${\rm H}$ represents the cross-entropy between two distributions. These two pre-trained branches help KSS-G generate more realistic samples by providing auxiliary losses.

\subsubsection{Attribute Feature Extraction and Data Reconstruction}
We extract attribute representations from seen samples, which are then utilized to generate new samples. For each $\bm{x}^i \in \bm{X}^s$, we employ a set of feature extractors ${\rm EXT}=\{{\rm EXT}_{1},{\rm EXT}_{2},\ldots, {\rm EXT}_{M}\}$ to extract a group of attribute features $\bm{g}^i = [\bm{g}^i_1, \ldots, \bm{g}^i_M] \in \mathbb{R}^{M \times \hat{d}}$, where $\hat{d}$ denotes the feature size. Each row of $\bm{g}^i$ corresponds to a distinct attribute in $\bm{z}^i$. In each batch, $B$ samples of each category $j$ are randomly selected from the training set. Subsequently, the attribute features $\bm{G}^j=[\bm{g}^{j} (1), \bm{g}^{j} (2), \ldots, \bm{g}^{j} (B)] \in \mathbb{R}^{B \times M \times \hat{d}}$, $j=1,\ldots, p$ are obtained from the ${\rm EXT}$ module. Additionally, a set of attribute recognizers ${\rm AR}=\{{\rm AR}_{1},{\rm AR}_{2},\ldots, {\rm AR}_{M}\}$ with a classification loss function $L_{att}$ is employed to ensure that attribute information is embedded in the features. 
\begin{equation}
\label{equa:lossatt}
    \mathcal{L}_{AR}=\frac{1}{N^s \times M}\sum_{i=1}^{N^s}\sum_{k=1}^{M} {\rm H} (z_{k}^{i},{\rm AR}_{k} ({\rm EXT}_{k}(\bm{x}^i))).
\end{equation}

It is worth mentioning that while ${\rm AR}$ can classify samples directly due to the one-to-one matching of attribute vectors and class labels, it may not work well for generalized zero-shot learning as ${\rm AR}$ may overfit to seen faults. To improve consistency among features that represent the same attribute, we impose a constraint on the variance of such features from different samples within each batch. This results in these features being numerically close to each other. 
As previously mentioned, we select $B$ samples of each category $j$ to construct each batch, so the batch size is $B \times p$. In the batch $b$, we divide the attribute feature set $\{ \bm{g}^{i}_k\in \mathbb{R}^{\hat{d}}, i = 1, \ldots, p \times B \}$ into two groups based on the value of $z^i_k$: $\{\bm{g}^i_k,z^i_k=1\}$ and $\{\bm{g}^i_k,z^i_k=0\}$ for each attribute $k$. We then calculate the variance of features in each group using the consistency loss function $L_{AV}$, which can be denoted as
\begin{equation}
    \mathcal{L}_{AV} \!=\! \frac{1}{N^b \!\times\! \hat{d}}\sum_{b=1}^{N^b}\sum^{\hat{d}}_{n=1}\sum_{k=1}^{M} (\sigma^n_b\{\bm{g}^i_k,z^i_k\!=\!1\} +\sigma^n_b\{\bm{g}^i_k,z^i_k\!=\!0\}),
\end{equation}
where $\sigma^n_b$ represents the variance of $\{\bm{g}^s_i|z^s_i=1 {\thinspace \rm or \thinspace} 0 \}$ in the $n$th dimension for the $b$th batch. $N^b$ denotes the number of batches determined by the number of training samples $N^s$ and the batch size $B \times p$. For each attribute $k$, this can be viewed as an operation to compact the distribution of features, which aids in attribute recognition. Additionally, the reconstruction module ${\rm REC}$ reconstructs the input $\bm{x}^i$ using attribute features to ensure that the original information is preserved. The corresponding reconstruction loss can be expressed as 
\begin{equation}
\label{equa:lossrec}
    \mathcal{L}_{R}=\frac{1}{N^s} \sum_{i=1}^{N^s}{\rm MSE} (\bm{x}^i, {\rm REC} (\bm{g}^i)),
\end{equation}
where ${\rm MSE}$ denotes the mean square error. 

\subsubsection{Generating Samples by Feature Group Reorganization}
The feature group reorganization operation is illustrated in Fig. \ref{fig:FGR}. During the training stage, KSS-G utilizes auxiliary knowledge to interpretably recombine attribute features and generate samples of seen faults in the following steps: 
\begin{enumerate}
\item[(i)]{\emph{Similar Category Search}: For each seen category $j$, we calculate the L1-norm distance between $\bm{a}^j$ and each row of $\bm{A}^s$ except for the row $j$ to find the seen category $m$ that is most similar to category $j$.}
\item[(ii)]{\emph{Feature Group Reorganization}: The fake samples of category $j$ are generated using the feature of category $m$. Specifically, we replace unsuitable parts of the attribute features extracted from samples of category $m$ by comparing $\bm{a}^j$ with $\bm{a}^m$, resulting in the fake feature of category $j$. For instance, if only one attribute differs between $\bm{a}^j$ and $\bm{a}^m$, such as $a^j_2=1$, $a^m_2=0$, for each $\bm{g}^{m} (i) = [\bm{g}^m_1 (i), \bm{g}^m_2 (i), \ldots, \bm{g}^m_M (i)]$ in $\bm{G}^m$, we randomly select a sample $h$ from those that satisfy $z^h_2=1$. We then use the corresponding feature $\bm{g}^{h}_2 (i)$ to replace the inappropriate feature $\bm{g}^{m}_{2} (i)$, resulting in the target feature $\widetilde{\bm{g}}^j (i) = [\bm{g}^m_1 (i), \bm{g}^{\textcolor[rgb]{1,0,0}{h}}_2 (i), \ldots, \bm{g}^m_M (i)]$. After the feature group reorganization operation, we get the $\widetilde{\bm{G}}^{j}$ based on $\bm{G}^m$. }
\item[(iii)]{\emph{Sample Generation}: We generate fake samples $\widetilde{\bm{x}}^i$ for each seen category $j$ using $\bm{\widetilde{G}}^{j}$ and the ${\rm REC}$.}
\end{enumerate}

\subsubsection{Auxiliary Loss from Aid-Discriminator}
The quality of fake samples is crucial in addressing DSP. In this regard, the discrimination branch of the aid-discriminator functions as a discriminator in GAN, which helps generate more realistic samples. This branch is represented by $\Theta_{d}$ and its loss can be expressed as follows:
\begin{equation}
    \mathcal{L}_{D} = -\mathbb{E}\left[log (D (\bm{x}^i))\right]-\mathbb{E}\left[log (1-D (\widetilde{\bm{x}}^i))\right].
\end{equation}

Correspondingly, the loss function of the generator can be expressed as
\begin{equation}
    \mathcal{L}_G = \mathbb{E}\left[log (1-D (\widetilde{\bm{x}}^i))\right].
\end{equation}

In addition to the discrimination branch, the multiclass classification and attribute recognition branches also contribute to generating more authentic samples in KSS-G. Therefore, the auxiliary loss can be expressed as follows:
\begin{equation}
\begin{split}
    \mathcal{L}_{AU} = \frac{1}{N^s}\sum_{i=1}^{N^s}\left [{\rm H} (y^i,D_y(\widetilde{\bm{x}}^i))+
    \frac{1}{M}\sum_{k=1}^{M} {\rm H} (z_{k}^i,\widetilde{z}^i_{k})\right ],
\end{split}
\end{equation}
where $\widetilde{\bm{z}}^i=D_a(\widetilde{\bm{x}}^i)$ and $\widetilde{z}^s_{k}$ is the $k$th element of $\widetilde{\bm{z}}^i$. It is worth noting that these two branches are not trained in this stage. The loss function of KSS-G can be denoted as
\begin{equation}
\begin{split}
    \mathcal{L}_{KSS-G}= \lambda_{AR} \times \mathcal{L}_{AR} +
    \lambda_{AV} \times \mathcal{L}_{AV} + \lambda_{AU} \times \mathcal{L}_{AU}\\
    + \lambda_{R} \times \mathcal{L}_{R} +
    \lambda_G \times \mathcal{L}_G,
\end{split}
\end{equation}
where $\lambda_{AR}$, $\lambda_{AV}$, $\lambda_{AU}$, $\lambda_{R}$, $\lambda_G$ are hyper parameters.

\subsection{Addressing Domain Bias via KSS-D}
\label{KSS-D}
\begin{figure}[!t]
    \centering
    \includegraphics[width=1 \linewidth]{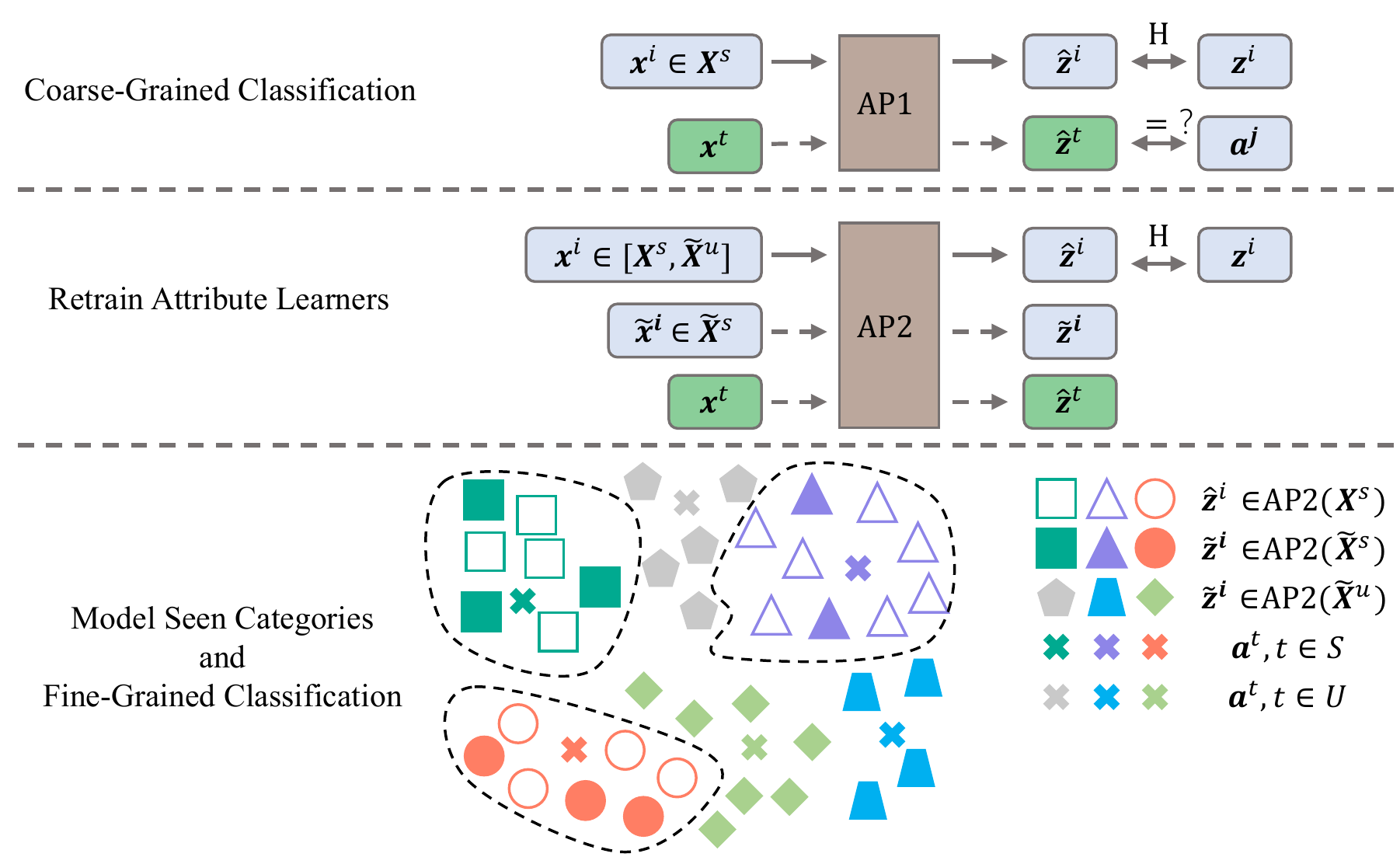}
    \caption{The diagram of the discrimination module (KSS-D). In the KSS-D, a projector trained by real and fake samples projects all samples into the knowledge space. Then control limits of seen categories are established based on real samples and fake samples to identify the projections of test samples in the knowledge space. }
    \label{fig:KSS-D}
\end{figure}

The KSS-D module is shown in Fig. \ref{fig:KSS-D}. To distinguish between samples belonging to seen or unseen categories, KSS-D projects the samples into the knowledge space and models the seen categories within that space. This approach allows us to determine the control limit of each seen category. If a test sample $\bm{x}^t \in \bm{X}^t$ does not belong to any seen category, it is classified as an unseen category, and a TZSD model, such as DAP, is trained to classify it. KSS-G generates fake samples of both seen and unseen categories to improve KSS-D. Overall, KSS-D provides an effective method for distinguishing between seen and unseen categories in the knowledge space while utilizing auxiliary knowledge and fake samples for improved accuracy. 
    
\subsubsection{Coarse-Grained Classification} 
Firstly, we train the attribute projector ${\rm AP}=\{{\rm AP}_1, {\rm AP}_2, \ldots,{\rm AP}_M\}$ to establish a mapping between $\bm{x}^i$ and $\bm{z}^i$. The attribute projector serves as a bridge between knowledge and samples, allowing for the integration of knowledge into the KSS-D. We denote this ${\rm AP}$ as ${\rm AP1}$.
\begin{equation}
    \mathcal{L}_{AP1}=\frac{1}{N^s \times M}\sum_{i=1}^{N^s}\sum_{k=1}^{M} {\rm H} (z_{k}^{i},\hat{z}_{k}^{i}),
\end{equation}
where $\hat{z}_{k}^{i}={\rm AP1}_{k} (\bm{x}^i)$, $\bm{x}^i \in \bm{X}^s$. After training, we use ${\rm AP1}$ to predict $\hat{\bm{z}}^t$ from $\bm{x}^t$ and normalize the output to either 1 or 0, depending on the confidence level, i.e., $\hat{\bm{z}}^t=\mathds{1} ({\rm AP1}(\bm{x}^t)>0.5)$. If $\hat{\bm{z}}^t$ is equal to any of the $\bm{a}^j \in \bm{A}^s$, then we determine the label of $\bm{x}^t$ be $y^j$. However, if none of the $\bm{a}^j$ are equal to $\hat{\bm{z}}^t$, we need to judge $\bm{x}^t$ using the following steps. 

\subsubsection{Retrain Attribute Learners} 
Similar to the training process of KSS-G, we use the similar category search to find the seen category $m$ that is most similar to each unseen category $j$. Then, we generate fake unseen samples by randomly selecting samples from category $m$ and reorganizing their feature groups. 
To maintain the balance between categories, KSS-G generates the same number of samples for each unseen category as for the seen categories in the training set.
We retrain the ${\rm AP}$ using real samples ${\bm{X}}^s$ and ${\bm{Z}}^s$, as well as fake unseen samples $\widetilde{\bm{X}}^u \in \mathbb{R}^{\hat{N}^u \times d}$ and corresponding attribute labels $\widetilde{{\bm{Z}}}^u$. This retrained ${\rm AP}$ is denoted as ${\rm AP2}$. 
\begin{equation}
    \mathcal{L}_{AP2}\!=\!\frac{1}{(N^s+\hat{N}^u) \!\times\! M}\sum_{i=1}^{N^s+\hat{N}^u}\!\sum_{k=1}^{M} {\rm H} (z_{k}^{i},\hat{z}_{k}^{i}),
\end{equation}
where $\hat{z}_{k}^{i}={\rm AP2}_{k} (\bm{x}^i)$, $\bm{x}^i \in [\bm{X}^s, \widetilde{\bm{X}}^u]$. After retraining, we generate $\hat{N}^s_j$ fake samples $\widetilde{\bm{X}}^j$ for each seen category $j$ using KSS-G. For all seen categories, the fake samples can be denoted as $\widetilde{\bm{X}}^s=\left [\widetilde{\bm{X}}^j|j=1,\cdots,p\right ] $. This generation process is similar to the one used for unseen category as described above. We then obtain $\widetilde{\bm{Z}}^j=[{\rm AP2}(\bm{X}^j),{\rm AP2}(\widetilde{\bm{X}}^j)] \in \mathbb{R}^{(N^s_j+\hat{N}^s_j) \times M}$ and $\hat{\bm{Z}}^t={\rm AP2}(\bm{X}^t) \in \mathbb{R}^{N^t \times M}$. Here, $N^s_j$ denotes the number of training samples of the seen category $j$. 

\subsubsection{Model Seen Categories} 
We model all seen categories in the knowledge space and build a control limit for each seen category. To model each seen category $j$, we employ a Gaussian Mixture Model (GMM) \cite{reynolds2009gaussian} and determine its control limit based on $\widetilde{\bm{Z}}^j$ in the knowledge space. The GMM for category $j$ can be represented as follows:
\begin{equation}
    {\rm P}^j(\widetilde{\bm{Z}}^j)=\sum_{r=1}^{R} \alpha_{r} {\rm N}\left (\widetilde{\bm{Z}}^j, \bm{\mu}_{r}, \bm{\sigma}_{r}\right),
\end{equation}
where $\bm{\mu}_{r} \in \mathbb{R}^{M}$ is the mean vector, $\bm{\sigma}_{r} \in \mathbb{R}^{M \times M}$ is the variance matrix, and $R$ is the component number, The coefficient of the $r$-th mixed component is denoted by $\alpha_{r}$. The function ${\rm N}\left (\widetilde{\bm{Z}}^j, \bm{\mu}_{r}, \bm{\sigma}_{r}\right)$ represents the Gaussian probability density function for the $r$-th component. 
\begin{equation}
\begin{split}
& {\rm N}\left (\widetilde{\bm{Z}}^j, \bm{\mu}_{r}, \bm{\sigma}_{r}\right)=\frac{1}{ (2 \pi)^{\frac{M}{2}}\left|\bm{\sigma}_{r}\right|^{\frac{1}{2}}}\\
&  \exp \left\{-\frac{1}{2}\left (\widetilde{\bm{Z}}^j-\bm{\mu}_{r}\right)^{T} \bm{\sigma}_{r}^{-1}\left (\widetilde{\bm{Z}}^j-\bm{\mu}_{r}\right)\right\}.
\end{split}
\end{equation}

The Expectation-Maximization algorithm \cite{do2008expectation} is utilized to fit the GMM. Once the model is fitted, we calculate the negative log-likelihood $\bm{I}^j$ or $\bm{I}^u$ for each sample $\bm{z}^i$ in $\widetilde{\bm{Z}}^j$ or $\bm{A}^u$. The calculation process for the $i$-th element in $\bm{I}^j$ or $\bm{I}^u$ can be expressed as follows:
\begin{equation}
    I_i = - \log\left ( \sum_{r=1}^{R} \alpha_{r} {\rm N} \left ( \bm{z}^i, \bm{\mu}_r, \bm{\sigma}_r  \right )  \right ).
\end{equation}

In this knowledge space, each category $j$ has a kernel point corresponding to $\bm{a}^j$, and samples should revolve around their corresponding kernel point in the knowledge space. To estimate the positions of unseen categories in the knowledge space to alleviate DSP, auxiliary knowledge $\bm{A}^u$ is used. The control limit for category $j$ is determined as 
\begin{equation}
    l^j= {\rm MIN}({\rm MAX}(\bm{I}^j),{\rm MIN} (\bm{I}^u)), 
\end{equation}

The control limit is designed to cover the space of seen categories as much as possible while avoiding coverage of unseen categories. This approach distinguishes between seen and unseen categories in the knowledge space, which helps to avoid the DSP because the distribution of each unseen category sample in the knowledge space should revolve around its attribute vector. We give different control limits for different seen categories for the following reasons. On the one hand, modeling each category separately allows for a more precise understanding of the feature distribution within each category. On the other hand, this approach can also address a more challenging scenario where a seen category $j$ and an unseen category are closely related in the semantic space. According to formula (15), only the control limit of the seen category $j$ will shrink in this case, while the control limits of other seen categories will remain unaffected. 

\subsubsection{Fine-Grained Classification}
We obtain the confidence vector (negative log-likelihood) $\bm{c}^t \in \mathbb{R}^p$ for the test sample $\bm{x}^t$, indicating its likelihood of belonging to each seen category based on its $\hat{\bm{z}}^t \in \hat{\bm{Z}}^t$ and GMM models, which can be expressed as
\begin{equation}
    c_j^t=-\log\left ({\rm P}^j(\hat{\bm{z}}^t)  \right ), j=1,2,\cdots,p.
\end{equation} 

If the value of element $c_j^t$ in $\bm{c}^t$ is greater than the corresponding $l^j$ for all seen categories, then $\bm{x}^t$ belongs to the unseen categories. Otherwise, it belongs to the seen categories $j$, where the L1-norm distance between $\bm{a}^j$ and $\hat{\bm{z}}^t$ is minimum. To achieve GZSD, a TZSD model is trained for unseen samples using fake unseen samples $\widetilde{\bm{X}}^u$ and real samples $\bm{X}^s$. The training stage of our proposed method is presented in Algorithm \ref{alg:gzsd}. The first part is about the training of the KSS-G, and the second part is about the training of the KSS-D. The testing stage is shown in Algorithm \ref{alg:online}. The first part is about coarse-grained classification and the second part is about fine-grained classification. 
\begin{algorithm}[!t]
	\renewcommand{\algorithmicrequire}{\textbf{Input:}}
	\renewcommand{\algorithmicensure}{\textbf{Output:}}
	\caption{The Training Stage of Generalized Zero-shot Diagnose via Knowledge Space Sharing Model}
	\label{alg:gzsd}
	\begin{algorithmic}[1]
		\REQUIRE The training dataset $\mathbb{D}_{tr}=\{ (\bm{x}^{i},y^{i},\bm{z}^{i})|\bm{x}^{i} \in \bm{X}^s,y^{i} \in \bm{Y}^s, \bm{z}^{i} \in \bm{Z}^s \}$, the fault category-attributes matrix $\bm{A} = [\bm{A}^s, \bm{A}^u]$.
		\ENSURE AP1, AP2, TZSD, GMM: $\{ {\rm P}^j, j=1,2,\cdots,p \}$, the control limit $\{l^j, j=1,2,\cdots,p\}$.
        \STATE $\slash *$  Train KSS-G  $* \slash$
        \FOR{$e=1,2,\cdots,E_p$}
        \STATE $\Theta_{b} \leftarrow \Theta_{b} - \epsilon \partial \mathcal{L}_{AD}(\Theta_{b};\mathbb{D}_{tr}) / \partial \Theta_{b}$;
        \ENDFOR
        \FOR{$e=1,2,\cdots,E$}
            \FOR{$b=1,2,\cdots,N^b$}
                \FOR{$w=1,2,\cdots,W$}
                    \STATE Randomly select B samples of each seen category $j$, i.e., $\{\bm{x}^i\}_{i=1}^{B \times p}$;
                    \STATE Generate B fake samples of each seen category $j$ by KSS-G, i.e., $\{\widetilde{\bm{x}}^i\}_{i=1}^{B \times p}$;
                    \STATE $\Theta_{d} \leftarrow \Theta_{d} - \epsilon \partial \mathcal{L}_{D}(\Theta_{d};\{\bm{x}^i\}_{i=1}^{B \times p}, \{\widetilde{\bm{x}}^i\}_{i=1}^{B \times p}) / \partial \Theta_{d}$;
                \ENDFOR
                \STATE Randomly select B samples of each seen category $j$, i.e., $\{\bm{x}^i\}_{i=1}^{B \times p}$;
                \STATE Generate B fake samples of each seen category $j$, i.e., $\{\widetilde{\bm{x}}^i\}_{i=1}^{B \times p}$;
                \STATE $\Theta_{g}\! \leftarrow\! \Theta_{g} \!- \epsilon \partial \mathcal{L}_{KSS-G}(\Theta_{g};\!\{\bm{x}^i\}_{i=1}^{B \times p}\!,\! \{\widetilde{\bm{x}}^i\}_{i=1}^{B \times p}) / \partial \Theta_{g}$;
            \ENDFOR
        \ENDFOR
        \STATE $\slash *$  Train KSS-D  $* \slash$
        \STATE Train ${\rm AP1}$ using $\bm{X}^s, \bm{Z}^{s}$ based on (10).
        \STATE Train ${\rm AP2}$ using $\bm{X}^s, \bm{Z}^{s}, \widetilde{\bm{X}}^u, \widetilde{{\bm{Z}}}^u$ based on formula (11).
        \STATE Use KSS-G to generate $\hat{N}^s_j$ fake samples $\widetilde{\bm{X}}^j$ for each seen category $j$. $\widetilde{\bm{Z}}^j=[{\rm AP2}(\bm{X}^j),{\rm AP2}(\widetilde{\bm{X}}^j)]$.
        \FOR{$j=1,2,\cdots,p$}
            \STATE Use GMM ${\rm P}^j$ to model each seen category $j$ based on $\widetilde{\bm{Z}}^j$ according to formula (12), (13);
            \STATE Calculate the control limit $l^j$ of category $j$;
        \ENDFOR
        \STATE Train TZSD model based on $\bm{X}^s, \bm{Z}^{s}, \widetilde{\bm{X}}^u, \widetilde{{\bm{Z}}}^u$.
	\end{algorithmic}  
\end{algorithm}
\begin{algorithm}[!t]
	\renewcommand{\algorithmicrequire}{\textbf{Input:}}
	\renewcommand{\algorithmicensure}{\textbf{Output:}}
	\caption{The Testing Stage of Generalized Zero-shot Diagnose via Knowledge Space Sharing Model}
	\label{alg:online}
	\begin{algorithmic}[1]
		\REQUIRE The testing samples $\bm{x}^{i} \in \bm{X}^t$, AP1, AP2, TZSD, GMM: $\{{\rm P}^j, j=1,2,\cdots,p\}$, the control limit $\{l^j, j=1,2,\cdots,p\}$.
		\ENSURE The classification results of testing samples.
        \FOR{each test sample $\bm{x}^i$}
            \STATE $\slash *$  Coarse-grained classification  $* \slash$
            \STATE $\hat{\bm{z}}^i=\mathds{1} ({\rm AP1}(\bm{x}^i)>0.5)$;
            \STATE $\hat{y}^{i}=NULL$, FLAG1 = False;
            \FOR{$j=1,2,\cdots,p$}
                \IF{$\hat{\bm{z}}^i == \bm{a}^j$}
                    \STATE $\hat{y}^{i} = j$, FLAG1 = True; Output $\hat{y}^{i}$;
                \ENDIF
            \ENDFOR
            \STATE $\slash *$  Fine-grained classification  $* \slash$
            \IF{FLAG1 == False}
                \STATE $\hat{\bm{z}}^i={\rm AP2}(\bm{x}^i)$, FLAG2 == False;
                \FOR{$j=1,2,\cdots,p$}
                    \STATE $c_j^i=-\log\left ({\rm P}^j(\hat{\bm{z}}^i))  \right )$, $L1^j = |\bm{a}^j-\hat{\bm{z}}^i|$;
                    \IF{$c_j^i <= l^j$}
                        \STATE FLAG2 == True;
                    \ENDIF
                \ENDFOR
                \IF{FLAG2 == False}
                    \STATE $\hat{y}^{i} = {\rm TZSD}(\bm{x}^i)$; Output $\hat{y}^{i}$;
                \ELSE
                    \STATE $\hat{y}^{i} = j$, where $j$ satisfy that $L1^j$ is minimum for $j=1,2,\cdots,p$; Output $\hat{y}^{i}$;
                \ENDIF
            \ENDIF
        \ENDFOR
	\end{algorithmic}  
\end{algorithm}

\section{Case Study}
\label{sec:case}
\subsection{Description of the Tennessee-Eastman Process}
\begin{figure}[!t]
\centering
\includegraphics[width=\linewidth]{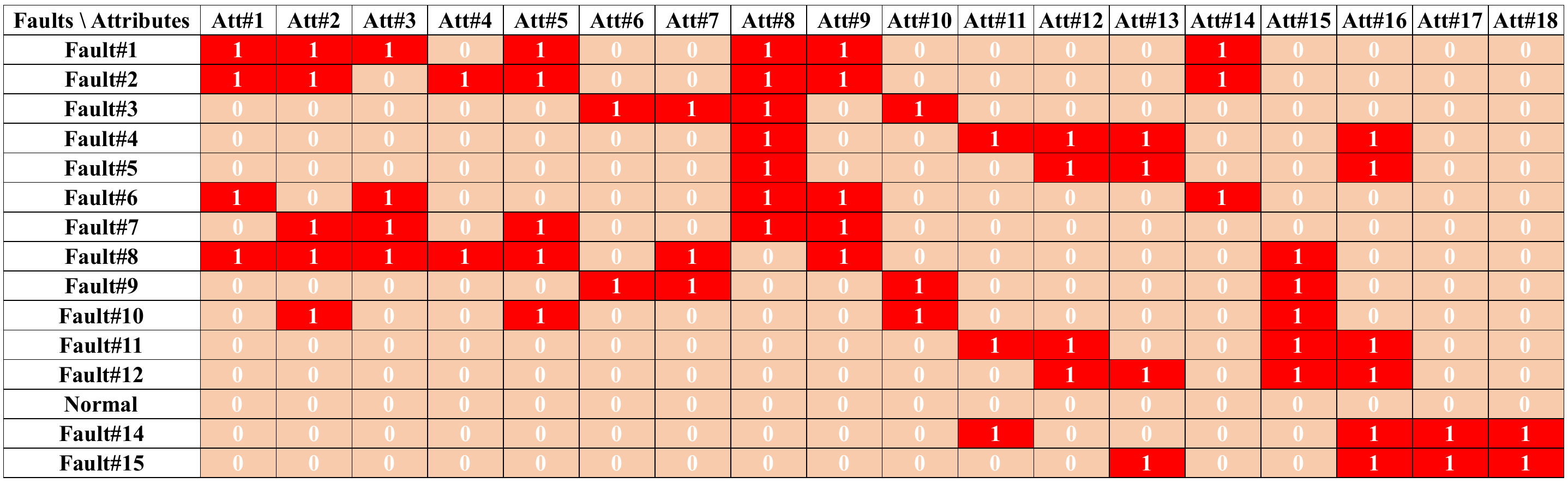}
\caption{Fault semantic attribute description. The "1" in the figure denotes the fault has this attribute, and the "0" denotes not.}
\label{fig:ATTmatrix}
\end{figure}

In this section, we explore the Tennessee-Eastman process (TEP) \cite{downs1993plant}, which is a widely used benchmark dataset for evaluating the performance of industrial models. TEP comprises five major subsystems, namely a reactor, a condenser, a vapor-liquid separator, a recycle compressor, and a product stripper. The dataset includes 41 measured variables and 11 manipulated variables, as well as 21 faults in addition to the normal working condition. For this case study, we select 14 types of faults along with the normal condition for training and testing purposes, which are described in Table \ref{tab:data_discription}. The fault semantic attributes for TEP are illustrated in Fig. \ref{fig:ATTmatrix}, and their specific names are listed in Table \ref{tab:att_discription}. 
\begin{table}[!t]
\centering
\caption{The Categories of TEP Used for The GZSL Fault Diagnosis}
\label{tab:data_discription}
\begin{tabular}{c|c|c}
\hline \hline No. & Fault state description & Disturb type \\
\hline 
1 & A/C feed ratio, B composition constant & Step change \\
2 & B composition, A/C ratio constant & Step change \\
3 & D feed temperature & Step change \\
4 & Reactor cooling water inlet temperature & Step change \\
5 & Condenser cooling water temperature & Step change \\
6 & A feed loss & Step change \\
7 & C header pressure loss & Step change \\
8 & A, B, C feed composition & Random variants \\
9 & D feed temperature & Random variants \\
10 & C feed temperature & Random variants \\
11 & Reactor cooling water inlet temperature & Random variants \\
12 & Condenser cooling water valve & Random variants \\
13 & Normal & - \\
14 & Reactor cooling water valve & Sticking \\
15 & Condenser cooling water valve & Sticking \\
\hline \hline
\end{tabular}
\end{table}%

\begin{table}[!t]
    \centering
    \caption{The Specific Information of Attributes for TEP}
    \begin{tabular}{c|c}
      \hline
      \hline
      No. & Attributes\\
      \hline
      Att\#1 & Input A is changed \\
      Att\#2 & Input C is changed \\
      Att\#3 & A/C ratio is changed \\
      Att\#4 & Input B is changed \\
      Att\#5 & Related with pipe4 \\
      Att\#6 & Temperature of input D is changed  \\
      Att\#7 & Related with pipe2 \\
      Att\#8 & Disturbance is step changing \\
      Att\#9 & Input is changed \\
      Att\#10 & Temperature of input is changed \\
      Att\#11 & Occurred at reactor \\
      Att\#12 & Temperature of cooling water is changed \\
      Att\#13 & Occurred at condenser \\
      Att\#14 & Related with pipe 1 \\
      Att\#15 & Disturbance is random varying \\
      Att\#16 & Related with cooling water \\
      Att\#17 & Related with valve \\
      Att\#18 & Disturbance is sticking\\
      \hline
      \hline
      \end{tabular}%
    \label{tab:att_discription}%
  \end{table}%

\subsection{Evaluation Setup and Model Details}
The 80\%-20\% split is used to divide the seen and unseen categories, resulting in 12 seen faults and 3 unseen faults. The division of these categories is randomly generated and presented in Table \ref{tab:data_division}. For each fault, 480 samples are collected for training and 800 for testing. Therefore, the number of training samples is 12*480, while the number of test samples is 15*800 (including 12*800 seen fault samples and 3*800 unseen fault samples). Table \ref{tab:details} provides details on the parameters of the knowledge space sharing model and the diagnosis procedure. The Leaky\_ReLU function serves as the activation function. The aid-discriminator undergoes pre-training for $E_p=100$ epochs, while KSS-G is trained for $E=300$ epochs. Additionally, wavelet denoising and z-score normalization are employed to preprocess data. To enhance training stability, the discrimination branch of aid-discriminator iterates $W$ times, while KSS-G updates once\cite{10.5555/2969033.2969125}. We set $W=2$ to balance performance and time trade-offs. Our goal is to achieve high accuracy on both seen and unseen categories; thus, we use the harmonic mean of classification accuracy for both categories to evaluate GZSL classification performance \cite{huang2021fault,Xian_2018_CVPR}. This evaluation can be expressed as:
\begin{equation}
    Har = 2 \times \frac{acc_s \times acc_u}{acc_s+acc_u},
\end{equation}
where $acc_s$ and $acc_u$ represent the average accuracy of seen and unseen faults, respectively. We choose the harmonic mean as our evaluation criteria, which is consistent with previous research \cite{zhuo2021auxiliary,huang2021fault}. The experiment results in the following are averaged diagnosis accuracies over five runs.
\begin{table}[!t]
    \centering
    \caption{The Five Groups of Train/Test Split For TEP}
    \label{tab:data_division}
    \begin{tabular}{c|c|c}
    \hline
    \hline
    No.   & Training Faults & Target Faults \\
    \hline
    A     &2-6, 8-14       &1, 7, 15  \\
    \hline
    B     &1, 3-9, 11, 12, 14, 15       &2, 10, 13  \\
    \hline
    C     &1, 2, 4, 5, 7-13, 15       &3, 6, 14  \\
    \hline
    D     &2-5, 7-9, 11-15       &1, 6, 10  \\
    \hline
    E     &1-3, 5-7, 9, 10, 12-15       &4, 8, 11  \\
    \hline
    \hline
    \end{tabular}%
\end{table}%
\begin{table*}[htbp]
  \centering
  \caption{Implementation Details}
  \begin{threeparttable}
    \begin{tabular}{cc|cc}
    \hline
    \hline
    \multicolumn{4}{c}{\textbf{KSS Model}}\\
    \hline
    \multicolumn{2}{c|}{\textbf{KSS-G}} & \multicolumn{2}{c}{\textbf{KSS-D}}\\
    \textbf{Feature extraction:} & \{FC (52 ($d$), 104) + FC (104, 32) + FC (32, 16)\} $\times$ 14 ($M$)\tnote{1} & \textbf{Class distribution:} & GMM$\times$12 ($p$)\\
    \textbf{Reconstruction:} & FC (16, 64) + Tran + FC (14 ($M$), 1) + Sque + FC (64, 52 ($d$)) \tnote{2} & \textbf{TZSD Model:} & DAP (Naive Bayes) \\
    \textbf{Attribute classification:} & \{FC (16, 32) + FC (32, 16) + FC (16, 2)\}$\times$14 ($M$) & $\rm AP$: & DAP (Random Forest)\\
    \hline
    \multicolumn{2}{c|}{\textbf{Aid-Discriminator}} & \multicolumn{2}{c}{\textbf{Training Process}}\\
    \textbf{Backbone:} & FC (52 ($d$), 256) + FC (256, 128) + FC (128, 64) & $\lambda_{R}=4, R=3$ &  \textbf{Optimizer:} AdamW\\
    \textbf{Multi-class classification:} & FC (64, 12 ($p$)) & $\lambda_{AR}=\lambda_{G}=1$ & $B=256$\\
    \textbf{Attribute recognition:} & FC (64, 28 ($M \times 2$)) & $\lambda_{AV}=\lambda_{AU}=0.5$ & $\hat{N}^s_j=N^s_j$\\
    \textbf{ Discrimination:} & FC (64, 16) + FC (16, 1) & Learning rate: $\epsilon = 0.001$ & $\hat{N}^u=\frac{N^s \times q}{p}$\\
    \hline
    \hline
    \end{tabular}%
  \label{tab:details}%

    \begin{tablenotes}
    \footnotesize
    \item[1] ‘+’ denotes that the input tensor passes through these layers in order. FC($a$, $b$) represents a fully connected layer with input dimension $a$ and output dimension $b$.
    \item[2] Tran denotes that the input tensor is transposed. Sque denotes that the last dimension of the tensor is compressed.
    \end{tablenotes}
\end{threeparttable}
\end{table*}%

\subsection{Results}
\subsubsection{Comparison of Accuracy with Other GZSL Methods}
We compare our method with three classic GZSL methods, namely ESZSL\cite{romera2015embarrassingly}, ALE\cite{akata2015label}, and SJE\cite{akata2015evaluation}. These embedding-based generalized zero-shot methods have been used as comparison methods in other research on ZSD \cite{feng2020fault,zhuo2021auxiliary}. Additionally, we compare our approach with GatingAE \cite{kwon2022gating}, which is one of the most recent generalized zero-shot learning methods that also employs a gating mechanism. Furthermore, we compare the proposed method with two latest GZSD methods, namely CVAE\cite{huang2021fault} and FAGAN\cite{zhuo2021auxiliary}. Although we have investigated a variety of other ZSD methods in Table \ref{tab:model_com_detail}, they are not suitable for comparison due to differences in application scenarios or settings during the test stage. 

Table \ref{tab:gzslresult} presents the accuracy (\%) of the methods evaluated. In the task of GZSD, test samples come from both seen and unseen categories, and it is important to comprehensively evaluate diagnostic performance for both. Our method outperforms the others in the harmonic mean index, which aligns with the GZSL task objective. Classic methods (ESZSL, ALE, SJE) perform poorly on the TE benchmark due to the lack of mechanisms that can improve generalization from seen to unseen categories and address DSP. Both FAGAN and CVAE are generative-based methods that perform well on seen categories but have unsatisfactory accuracies on unseen faults. FAGAN lacks a mechanism to distinguish seen and unseen categories, while CVAE's binary classifier trained on seen samples and fake unseen samples performs poorly due to difficulty in generating high-quality unseen samples. Compared to Gating AE, our gating mechanism is based on knowledge space where an unseen class's position can be directly determined by its auxiliary knowledge, avoiding inaccurate estimation of its prototype. Additionally, our gating mechanism is based on absolute distances from test samples to distributions of each seen class rather than relative distances. Our approach is more suitable for GZSD tasks as it can accurately detect unseen class samples that perform differently from seen classes.
\begin{table*}[htbp]
\centering
\caption{Results of Generalized Zero-Shot Diagnosis on TEP}
    \begin{tabular}{c|c|c|c|c|c|c|c|c|c|c|c|c|c|c|c}
    \hline
    \hline
    \multirow{2}{*}{Methods}     & \multicolumn{3}{c|}{A} & \multicolumn{3}{c|}{B} & \multicolumn{3}{c|}{C} & \multicolumn{3}{c|}{D} & \multicolumn{3}{c}{E} \\
\cline{2-16}          & $acc_s$    & $acc_u$    & $Har$     & $acc_s$    & $acc_u$    & $Har$     & $acc_s$    & $acc_u$    & $Har$     & $acc_s$    & $acc_u$    & $Har$     & $acc_s$    & $acc_u$    & $Har$ \\
    \hline
    ESZSL & 8.63     & 7.04     & 7.75     & 5.71     & 18.00     & 8.67     & 6.82     & 25.17     & 10.74     & 4.53    & 26.58    & 7.74    & 7.09    & 5.71    & 6.33 \\
    ALE   & 7.77     & 6.04     & 6.80     & 6.42     & 8.88     & 7.45     & 7.06     & 21.06     & 10.57     & 5.45    & 25.11    & 8.94    & 11.12    & 7.74    & 9.07 \\
    SJE   & 8.42     & 0.00     & 0.00     & 7.19     & 7.75     & 7.46     & 2.41     & 42.08     & 4.55     & 8.39    & 15.49    & 10.85    & 8.57    & 0.00    & 0.00 \\
    GatingAE   & 21.64 & 20.83   & 21.23     & 23.76     & 36.25     & 28.71     & 61.10     & 1.15     & 2.25     & 35.15    & 33.00    & 34.04    & 48.52    & 5.48    & 9.85 \\
    CVAE  & \bf{59.13}     & 24.97     & 34.44     & \bf{69.36}     & 30.10     & 41.93     & \bf{72.99}     & 12.77     & 21.73     & 61.77    & 29.83    & 40.19    & 57.54    & 0.75    & 1.48 \\
    FAGAN & 53.62     & 0.00     & 0.00     & 61.83     & 0.00     & 0.00     & 63.27     & 2.13     & 4.11     & \bf{61.94}    & 0.17    & 0.33    & \bf{58.42}    & 0.46    & 0.91 \\
    \textbf{KSS (ours)} & 50.30     & \bf{48.61}     & \bf{49.44}     & 54.98     & \bf{49.33}     & \bf{51.99}     & 58.98     & \bf{42.33}     & \bf{49.29}     & 50.72    & \bf{70.54}    & \bf{59.01}    & 51.41    & \bf{33.58}    & \bf{40.63}
    \\
    \hline
    \hline
    \end{tabular}%
\label{tab:gzslresult}%
\end{table*}

\subsubsection{Visualization Study}
\begin{figure}[!t]
    \begin{minipage}[t]{0.5\linewidth}
        \centering
        \includegraphics[width=0.94\textwidth]{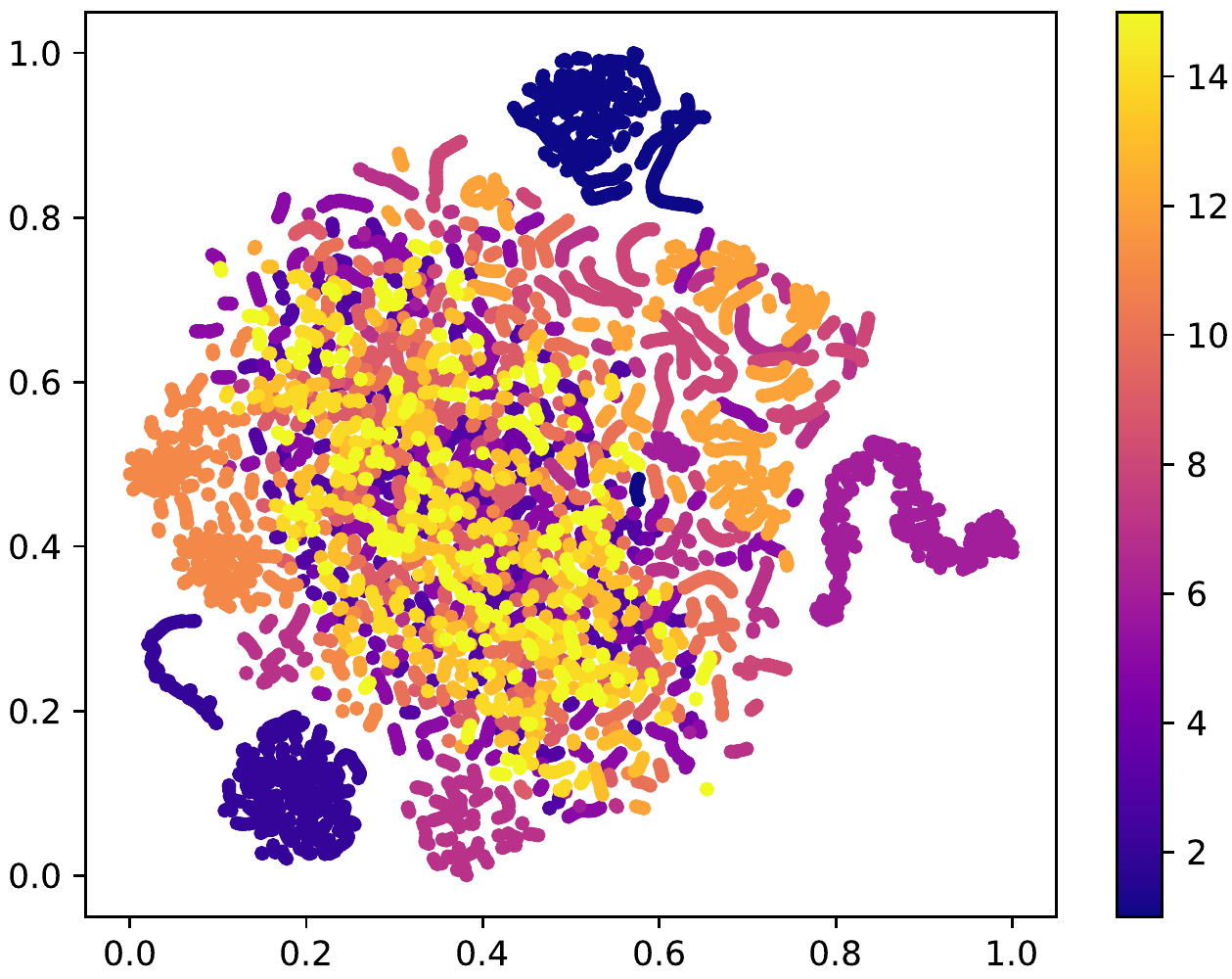}
        \centerline{ (a)}
    \end{minipage}%
    \begin{minipage}[t]{0.5\linewidth}
        \centering
        \includegraphics[width=\textwidth]{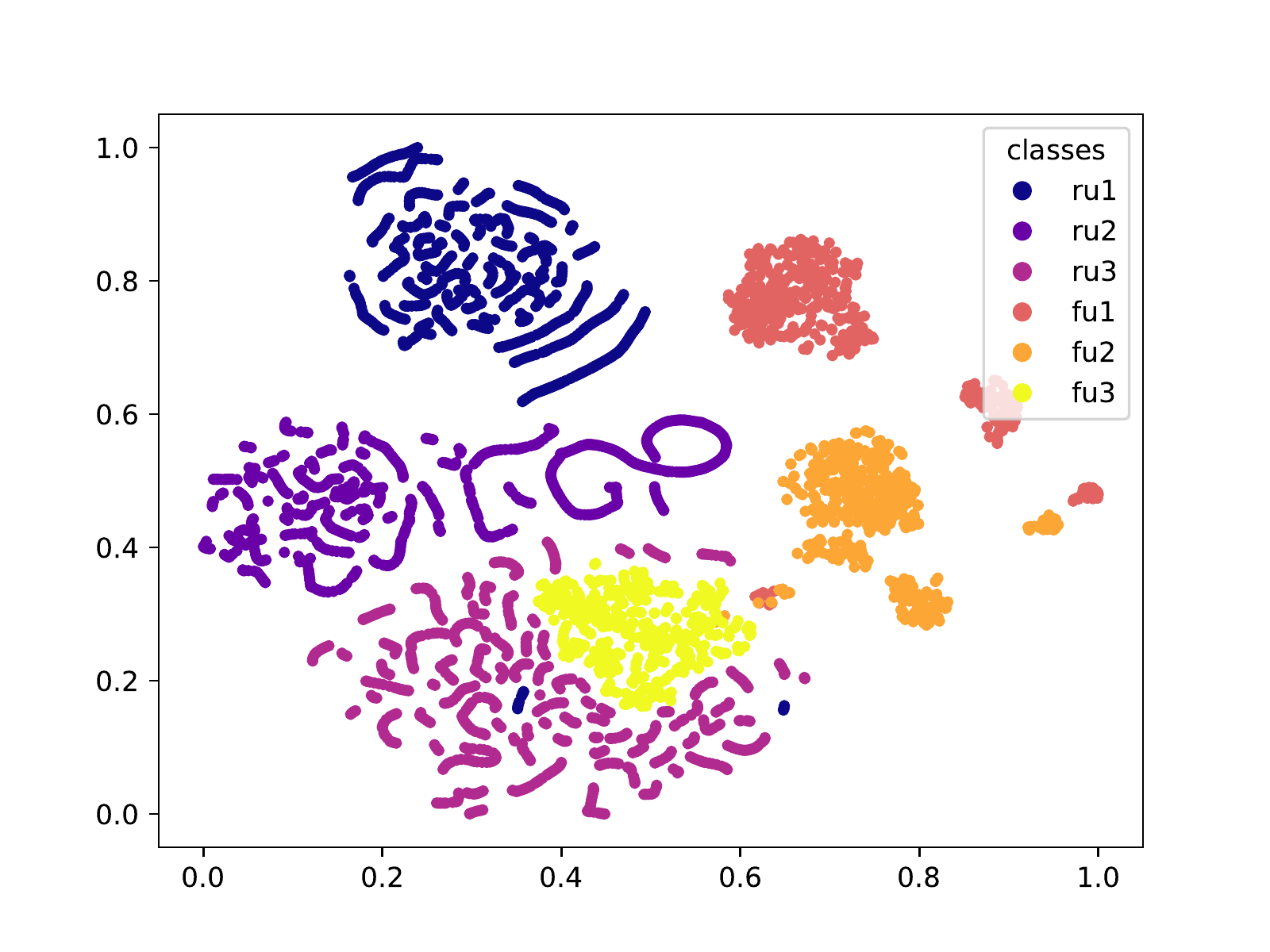}
        \centerline{ (b)}
    \end{minipage}
    \caption{(a) T-SNE visualization of test samples. (b) T-SNE visualization of test and generated samples of unseen categories (Group A). ru1, ru2, and ru3 correspond to the real samples of unseen categories. Also, fu1, fu2, and fu3 correspond to the fake samples of unseen categories.}
    \label{fig:T-SNEall}
\end{figure}

Figure \ref{fig:T-SNEall} (a) displays the distribution of all test samples. The industrial fault diagnosis task is challenging, partly due to the insignificant difference between samples of different categories. Figure \ref{fig:T-SNEall} (b) illustrates the samples of unseen categories in the test dataset and generated samples of unseen categories. The figure suggests that the distribution of generated samples closely resembles the actual distribution, which aids in distinguishing between seen and unseen samples for KSS-D and enhances the performance of the TZSD model.

\subsubsection{Ablation Study}
Table \ref{tab:Ablation Study} illustrates the impact of loss functions on the proposed method's ability to perform GZSD. The experiments demonstrate that even without KSS-G, the proposed method can still perform GZSD to a certain extent. However, if we only rely on basic constraints to guide the generative process (No.2), low-quality samples can degrade performance by providing inaccurate information to KSS-D and the TZSD model. To address this issue, introducing an aid-discriminator to form a GAN structure (No.3) can supervise fake seen samples and improve their quality. Moreover, pre-training the aid-discriminator can enhance its discrimination ability (No.4). Additionally, using pre-trained multiclass classification and attribute recognition branches (No.5) can further improve KSS-G's generation quality. Finally, constraining the difference between features representing the same attribute value from different categories (No.6) makes attribute feature reorganization more reliable, thereby increasing performance. 
\begin{table}[!t]
  \centering
  \caption{Ablation Study}
    \begin{tabular}{cccc}
    \hline
    \hline
    Group & No. & Method & $Har$ (\%) \\
    \hline
    \multirow{6}{*}{A}  
        & 1  & w/o KSS-G & 41.58 \\
        & 2  & ${\rm No. 1}+\mathcal{L}_{AR}+\mathcal{L}_{R}$     & 33.13 (-20.32\%)  \\   
        & 3  & ${\rm No. 2} +\mathcal{L}_{G}$     & 37.42 (-10.04\%)  \\  
        & 4  & ${\rm No. 3}+\mathcal{L}_{AD}$ & 43.92 (+5.63\%) \\ 
        & 5  & ${\rm No. 4}+\mathcal{L}_{AU}$ & 45.82 (+10.20\%) \\ 
        & 6  & ${\rm No. 5}+\mathcal{L}_{AV}$    & 49.44 (+18.81\%) \\   
    \hline
    \multirow{6}{*}{C} 
        & 1  & w/o KSS-G & 34.66 \\
        & 2  & ${\rm No. 1}+\mathcal{L}_{AR}+\mathcal{L}_{R}$     & 25.82 (-25.50\%)  \\   
        & 3  & ${\rm No. 2} +\mathcal{L}_{G}$     & 37.89 (+9.32\%)  \\  
        & 4  & ${\rm No. 3}+\mathcal{L}_{AD}$ & 42.03 (+21.26\%) \\ 
        & 5  & ${\rm No. 4}+\mathcal{L}_{AU}$ & 46.52 (+34.22\%) \\ 
        & 6  & ${\rm No. 5}+\mathcal{L}_{AV}$    & 49.29 (+42.21\%) \\   
    \hline
    \hline
    \end{tabular}
  \label{tab:Ablation Study}
\end{table}

\subsubsection{Comparison with Generalized Few-shot Diagnosis}
For comparison, we conduct generalized few-shot diagnosis (GFZD) experiments. Specifically, we use 10, 50, 100, 200, or 480 samples of each unseen class and the same training samples of seen classes as GZSD experiments to train the methods used for comparison. The test data are the same as those used in GZSD. The compared algorithms include Support Vector Machine (linear), Logistic Regression (LR), Support Vector Machine (rbf), Random Forest (RF), and GaussianNB (NB). All these methods are implemented using the scikit-learn package \cite{pedregosa2011scikit}, and we uniformly adapt the default settings. As shown in Table \ref{tab:fewshot A} and Table \ref{tab:fewshot C}, conventional machine learning algorithms perform poorly when only a few unseen fault samples are provided for training. This is because they suffer from the lack of unseen samples in the GFZD task. Even for the supervised situation where all training samples of seen and unseen categories are provided, classification performance is still not satisfactory due to insignificant differences between different categories (as shown in Fig. \ref{fig:T-SNEall} (a)). The compared algorithms require at least 100 or 200 shots of unseen samples to achieve competitive results similar to ours. Our experimental results indicate that the concluded fault attributes indeed offer additional fault information for diagnosis purposes. Actually, it is unfair for the proposed methods to compare with the generalized few-shot diagnosis because compared with the few-shot setting, the zero-shot setting is more difficult.
\begin{table}[!t]
    \centering
    \caption{Comparison with Generalized Few-Shot Learning (Group A)}
    \begin{tabular}{c|c|c|c|c|c}
        \hline \hline \multirow{2}{*}{Group A: $Har$ (\%) } & \multicolumn{5}{|c}{Number of samples for each few-shot fault} \\
        \cline { 2 - 6 } & $\mathbf{10}$ & $\mathbf{50}$ & $\mathbf{100}$ & $\mathbf{2 0 0}$ & $\mathbf{480}$ \\
        \hline SVM (linear) & $4.82$ & $13.31$ & $52.15$ & $54.62$ & $55.09$ \\
        LR & $1.07$ & $7.47$ & $40.05$ & $52.42$ & $53.21$ \\
        SVM (rbf) & $0.66$ & $17.33$ & $42.54$ & $47.37$ & $59.79$ \\
        RF & $1.24$ & $7.61$ & $27.22$ & $62.39$ & $67.04$ \\
        NB & $0$ & $35.27$ & $58.09$ & $60.85$ & $59.30$ \\
        \hline Ours & \multicolumn{5}{|c}{$\mathbf{49.44}$} \\
    \hline \hline
    \end{tabular}
    \label{tab:fewshot A}
\end{table}
\begin{table}[!t]
    \centering
    \caption{Comparison with Generalized Few-Shot Learning (Group C)}
    \begin{tabular}{c|c|c|c|c|c}
        \hline \hline \multirow{2}{*}{Group C: $Har$ (\%) } & \multicolumn{5}{|c}{Number of samples for each few-shot fault} \\
        \cline { 2 - 6 } & $\mathbf{10}$ & $\mathbf{50}$ & $\mathbf{100}$ & $\mathbf{2 0 0}$ & $\mathbf{480}$ \\
        \hline SVM (linear) & $4.85$ & $13.83$ & $16.04$ & $24.29$ & $40.25$ \\
        LR & $2.28$ & $5.83$ & $12.78$ & $18.26$ & $38.69$ \\
        SVM (rbf) & $0.17$ & $4.82$ & $10.31$ & $44.34$ & $52.56$ \\
        RF & $0.33$ & $43.57$ & $45.79$ & $48.88$ & $57.94$ \\
        NB & $0$ & $8.83$ & $22.18$ & $47.67$ & $48.28$ \\
        \hline Ours & \multicolumn{5}{|c}{$\mathbf{49.29}$} \\
    \hline \hline
    \end{tabular}
    \label{tab:fewshot C}
\end{table}

\subsubsection{Exteneded TEP Dataset}
In this section, we re-evaluated our method on an extended TE dataset\cite{rieth2018issues} due to the relatively small amount of data provided by the classic original TE dataset for each fault. The extended version offers five hundred rounds of simulation for each fault, with random seeds varying from round to round. We selected the first five rounds of the training subset as the training set and the first five rounds of the testing subset as the testing set. We collected 480 * 5 samples of each seen fault for training and 800 * 5 samples of each fault for testing. The remaining experimental settings are kept consistent with those used for the original dataset. 

The experiment results are presented in Table \ref{tab:bigtep}. Based on the results, our method still achieves relatively high GZSD accuracy on the extended TEP dataset. When comparing the results of Table \ref{tab:gzslresult} and Table \ref{tab:bigtep}, it is evident that ESZSL, ALE, and SJE have limited ability to fit complex distributions due to their simple structure and fewer parameters. As a result, they do not improve accuracy on seen categories. On the other hand, other methods show an increase in accuracy on seen categories. This is expected since the training data for seen categories has increased in both quantity and diversity. However, nearly all methods exhibit a decrease in detection performance on unseen categories. This could be attributed to two factors: firstly, the unseen category data in the test set comes from multiple different simulations and is more diverse, which means the task is more difficult; secondly, generating both real and diverse data for unseen categories for these generative-based methods is more challenging.
\begin{table}[!t]
    \centering
    \caption{Results of Generalized Zero-Shot Diagnosis on Extended TEP}
      \begin{tabular}{c|c|c|c|c|c|c}
      \hline
      \hline
      \multirow{2}{*}{Methods} & \multicolumn{3}{c|}{A} & \multicolumn{3}{c}{C}\\
  \cline{2-7}          & $acc_s$   & $acc_u$   & $Har$     & $acc_s$   & $acc_u$   & $Har$\\
      \hline
      ESZSL & 5.89  & 4.44  & 5.06  & 9.24 & 14.42 & 11.27\\
      ALE   & 4.92  & 4.08  & 4.17  & 7.48  & 10.26 & 8.64 \\
      SJE   & 8.33  & 0.03  & 0.05  & 0.02  & \textbf{40.92} & 0.03 \\
      GatingAE   & 25.50  & 33.43  & 28.93  & 65.54  & 18.18 & 28.47 \\
      CVAE  & \textbf{72.43} & 0.00     & 0.00     & \textbf{76.78} & 18.58  & 29.90 \\
      FAGAN & 63.97 & 0.00  & 1.16  & 66.99  & 0.10     & 0.20 \\
      \textbf{KSS(ours)} & 62.80 & \textbf{36.78} & \textbf{46.39} & 70.44 & 36.18 & \textbf{47.24}\\
      \hline
      \hline
      \end{tabular}%
      \label{tab:bigtep}
  \end{table}%

\section{Conclusion}
\label{sec:con}
In this article, we propose a knowledge space sharing method to transfer knowledge between seen and unseen categories, which can be applied to address the domain shift problem. Compared to other methods for generalized zero-shot industrial fault diagnosis, on the one hand, our KSS-D method alleviates the DSP by only modeling the seen categories in the knowledge space with the help of auxiliary knowledge. The distribution shift problem between the training and testing set for seen categories refers to the research of transfer learning, which can be an interesting challenge for the GZSD task in the future. On the other hand, our proposed KSS-G method generates fake samples in an interpretable manner by extracting and recombining transferable attribute representations under the guidance of knowledge. The fake seen and unseen samples provided by KSS-G assist KSS-D in modeling and classifying seen categories, thereby relieving DSP. We have validated the excellent performance of our method for generalized zero-shot industrial fault diagnosis compared to SOTA methods on the benchmark dataset. In comparison with the generalized few-shot diagnosis, the advantage that zero sample of target faults is required can be observed. Even with no training samples for unseen faults, our method performs close to other methods that use 100 or 200 shots of each unseen fault.

\bibliographystyle{IEEEtran}  
\bibliography{IEEEabrv, tii-articles-template.bib}

\begin{IEEEbiography}[{\includegraphics[width=1in,height=1.25in,clip,keepaspectratio]{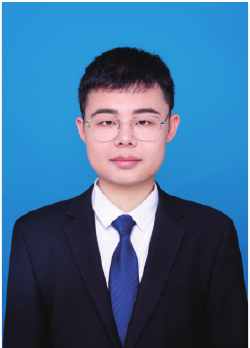}}]{Jiancheng Zhao} received B.Eng. degree in automation from College of Control Science and Engineering, Zhejiang University, Hangzhou, China, in 2021, where he is currently pursuing the Ph.D. degree. His current research interests include industrial big data, zero-shot learning, few-shot learning.
\end{IEEEbiography}

\begin{IEEEbiography}[{\includegraphics[width=1in,height=1.25in,clip,keepaspectratio]{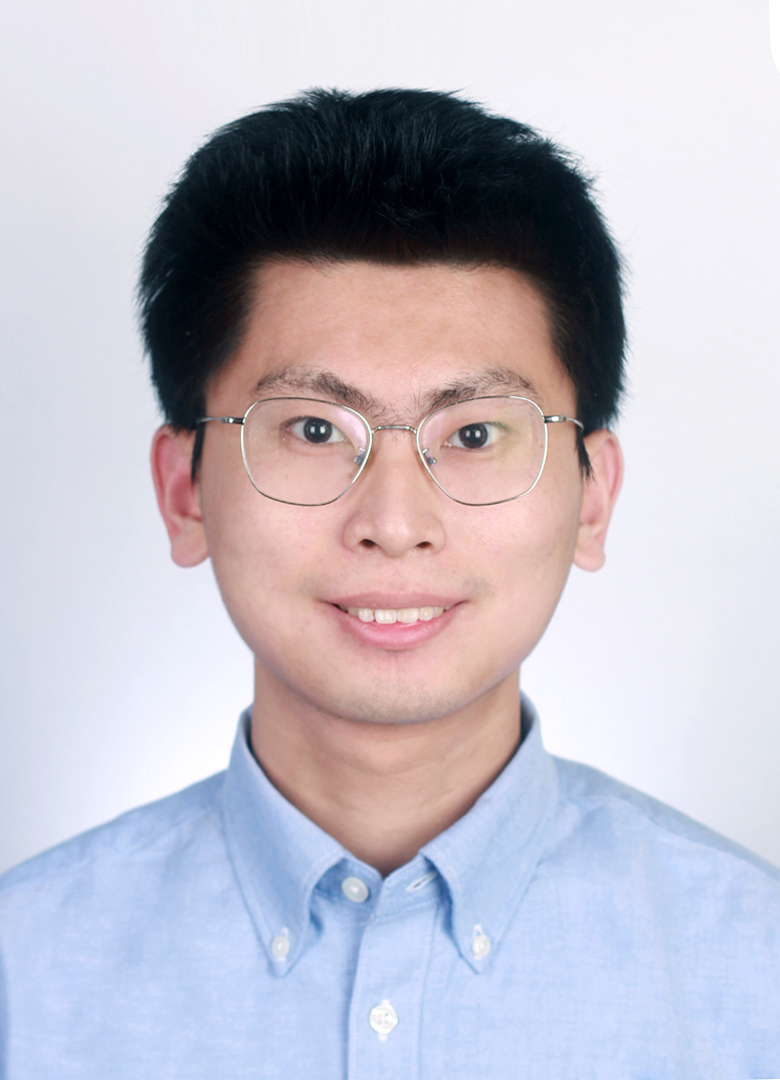}}]{Jiaqi Yue} received the B.Eng. degree in automation from the College of Electrical Engineering, Zhejiang University, Hangzhou, China, in 2022, where he is currently pursuing the Ph.D. degree in control science and engineering with the College of Control Science and Engineering. His current research interests include fault diagnosis and zero-shot learning.
\end{IEEEbiography}

\begin{IEEEbiography}[{\includegraphics[width=1in,height=1.25in, clip,keepaspectratio]{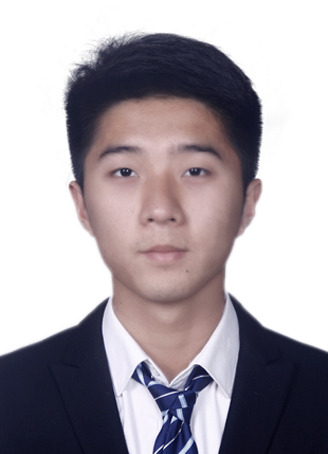}}]{Liangjun Feng} received the B.Eng. degree from North China Electric Power University, Beijing, China, in 2017. During the time he researched in the laboratory of Guotian Yang, studied as an exchange student in University of Wisconsin-Milwaukee, and got the Beijing outstanding graduate honor. Now, he is currently pursuing the Ph.D degree with the College of Control Science and Engineering, Zhejiang University, Hang-Zhou, China. His current research interests include machine learning, artificial intelligence, and pattern recognition.
\end{IEEEbiography}

\begin{IEEEbiography}[{\includegraphics[width=1in,height=1.25in,clip,keepaspectratio]{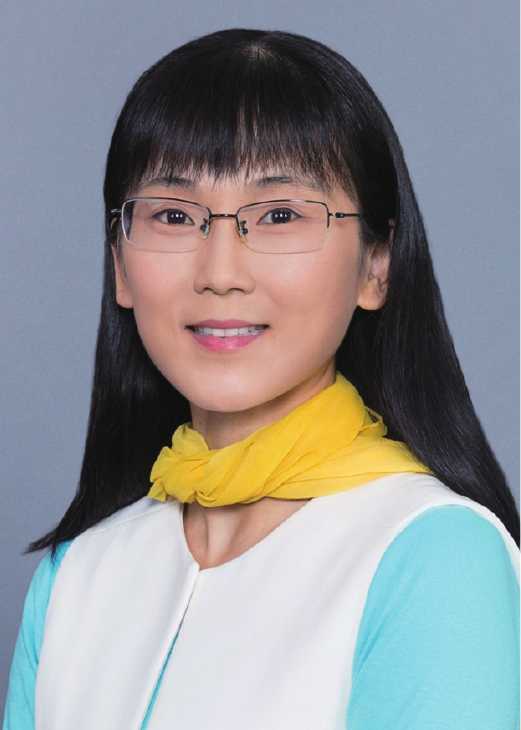}}]{Chunhui Zhao}
(SM'15) received Ph.D. degree from Northeastern University, China, in 2009. From 2009 to 2012, she was a Postdoctoral Fellow with the Hong Kong University of Science and Technology and the University of California, Santa Barbara, Los Angeles, CA, USA. Since January 2012, she has been a Professor with the College of Control Science and Engineering, Zhejiang University, Hangzhou, China. Her research interests include statistical machine learning and data mining for industrial application. She has authored or coauthored more than 140 papers in peer-reviewed international journals. She has served Senior Editor of Journal of Process Control, AEs of two International Journals, including Control Engineering Practice and Neurocomputing. 

She was the recipient of the National Top 100 Excellent Doctor Thesis Nomination Award, New Century Excellent Talents in University, China, and the National Science Fund for Excellent Young Scholars, respectively.
\end{IEEEbiography}

\begin{IEEEbiography}[{\includegraphics[width=1in,height=1.25in,clip,keepaspectratio]{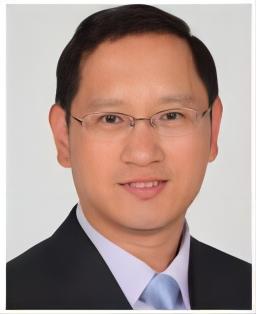}}]{Jinliang Ding}(Senior Member, IEEE) received the bachelor’s, master’s, and Ph.D. degrees in control theory and control engineering from Northeastern University, Shenyang, China, in 2001, 2004, and
2012, respectively. 

He is currently a Professor with the State Key Laboratory of Synthetical Automation for Process Industry, Northeastern University. He has authored or co-authored over 100 refereed journal articles and refereed papers at international conferences. He has also invented or co-invented 20 patents. His current research interests include modeling, plant-wide control, optimization for complex industrial systems, stochastic distribution control, and multiobjective evolutionary algorithms and their applications. 

Dr. Ding was a recipient of the three First-Prize of Science and Technology
Awards of the Ministry of Education in 2006, 2012, and 2018, respectively, the International Federation of Automatic Control (IFAC) Control Engineering
Practice for 2011–2013 Paper Prize, the National Technological Invention Award in 2013, the National Science Fund for Distinguished Young Scholars
in 2015, and the Young Scholars Science and Technology Award of China in 2016. One of his articles published on control engineering practice was selected 
for the Best Paper Award from 2011 to 2013.
\end{IEEEbiography}

\vfill

\end{document}